\documentclass{article}

\usepackage[numbers, sort&compress]{natbib}  

    \usepackage[preprint]{neurips_2025}

\usepackage[utf8]{inputenc} 
\usepackage[T1]{fontenc}    
\usepackage{hyperref}       
\usepackage{url}            
\usepackage{booktabs}       
\usepackage{amsfonts}       
\usepackage{nicefrac}       
\usepackage{microtype}      

\usepackage{graphicx}
\usepackage{multirow}
\usepackage{multicol}
\usepackage[table]{xcolor}
\usepackage{float}
\usepackage[countmax]{subfloat}
\usepackage{subcaption}
\usepackage{wrapfig}

\title{MLEP: Multi-granularity Local Entropy Patterns for \\Generalized AI-generated Image Detection}

\author{%
  Lin Yuan$^{1}$, Xiaowan Li$^{2}$, Yan Zhang$^{3}$, Jiawei Zhang$^{4}$, Hongbo Li$^{5}$, Xinbo Gao$^{6}$ \\
  $^{1-6}$Chongqing University of Posts and Telecommunications \\
  Chongqing 400065, China \\
  \texttt{yuanlin@cqupt.edu.cn} \\
  \texttt{lixiaowan.work@qq.com},  \\
  \texttt{\{yanzhang1991, zhangjw, lihongbo, gaoxb\}@cqupt.edu.cn}
}

\begin{document}

\maketitle

\begin{abstract} \label{sec:abstract}

Advancements in image generation technologies have raised significant concerns about their potential misuse, such as producing misinformation and deepfakes. Therefore, there is an urgent need for effective methods to detect AI-generated images (AIGI). Despite progress in AIGI detection, achieving reliable performance across diverse generation models and scenes remains challenging due to the lack of source-invariant features and limited generalization capabilities in existing methods. In this work, we explore the potential of using image entropy as a cue for AIGI detection and propose Multi-granularity Local Entropy Patterns (MLEP), a set of entropy feature maps computed across shuffled small patches over multiple image scaled. MLEP comprehensively captures pixel relationships across dimensions and scales while significantly disrupting image semantics, reducing potential content bias. Leveraging MLEP, a robust CNN-based classifier for AIGI detection can be trained. Extensive experiments conducted in an open-world scenario, evaluating images synthesized by 32 distinct generative models, demonstrate significant improvements over state-of-the-art methods in both accuracy and generalization.

\end{abstract}

\section{Introduction} \label{sec:introduction}

The rapid development of generative technologies has transformed image synthesis, with models like GAN~\cite{goodfellow2014gan}, diffusion model~\cite{ho2020denoising}, and their variants achieving impressive realism. While enabling new applications in creative industries, these advancements have also raised concerns over misuse in misinformation and deepfakes~\cite{thompson2023isn,xu2023misinfo}, prompting an urgent need for reliable AI-generated image (AIGI) detection methods. Researchers have leveraged spatial~\cite{wang2020cnn,zhong2024patchcraft,tan2024rethinking,zheng2024breaking} and frequency-domain cues~\cite{frank2020leveraging,qian2020thinking,luo2021generalizing,liu2022detecting}, as well as high-level knowledge from pretrained diffusion models~\cite{wang2023dire,chen2024drct} and LLMs~\cite{ojha2023towards,khan2024clipping,liu2024forgery} for AIGI detection. Yet, the lack of source-invariant representations still limits the cross-domain detection robustness, especially when across different models and content types.


To address this challenge, we aim to identify a generalized, content-agnostic pattern that can reliably distinguish AIGIs from real photographs. Inspired by recent studies~\cite{tan2024rethinking,zheng2024breaking}, our work builds on two key observations. Tan et al.\cite{tan2024rethinking} found that generative models typically involve internal upsampling operations and propose Neighboring Pixel Relationships (NPR) to capture resulting structural artifacts. However, NPR operates on small local patches and retains visible semantic structures, introducing bias that may hinder generalization. Zheng et al.\cite{zheng2024breaking} emphasized the impact of ``semantic artifacts'' on detection and propose disrupting image semantics by shuffling $32\times32$ patches. While this improves cross-scene generalization, the relatively large patch size still preserves semantic information. We argue that such artifacts persist and continue to limit content-agnostic detection.

\begin{figure}[t]
    \centering
    \includegraphics[width=\textwidth]{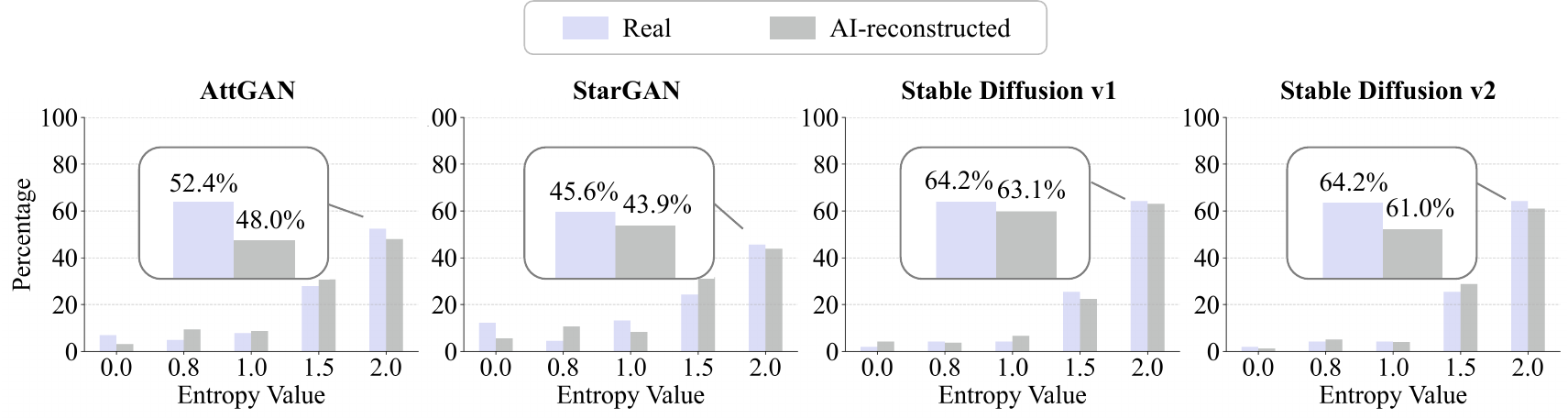}
    \caption{Comparison of local entropy distributions between real and AI-generated images using $2 \times 2$ patches, with entropy values from $\{0, 0.8, 1.0, 1.5, 2.0\}$. Real images consistently show a higher likelihood of entropy reaching 2.0.}
    \label{fig:intial_finding_entropy}
\end{figure}


Through large-scale subjective observation, we noticed a distinct ``glossy and smooth'' texture in AI-generated images, prompting an investigation into their entropy characteristics, a statistical measure of pixel randomness~\cite{shannon1948mathematical}. We conducted a preliminary study comparing local entropy distributions (using $2 \times 2$ patches) between real and AI-generated images. As shown in Fig.~\ref{fig:intial_finding_entropy}, real images consistently exhibit a higher probability of maximum entropy (2.0), suggesting the potential of entropy as a discriminative feature for AIGI detection. 
Motivated by this, we propose using image entropy as an alternative to pixel differences as proposed by NPR~\cite{tan2024rethinking}. Entropy captures pixel relationships while reducing semantic dependency by focusing on pixel value distributions rather than contrasts. To further suppress semantic artifacts, we adopt fine-grained patch shuffling (smaller than the $32 \times 32$ patches used in~\cite{zheng2024breaking}), which also reduces the computational overhead of entropy computation. Additionally, we incorporate multi-scale resampling and an overlapping sliding window to enhance the granularity of entropy patterns. Our contributions are summarized as follows:


\begin{itemize}
    \item To the best of our knowledge, it is the first attempt to explore the potential of 
    image entropy as a cue for detecting AI-generated images. Using image entropy not only 
    enhances detection accuracy and generalization compared to state-of-the-art 
    methods but also highlights intrinsic differences between real and AI-generated 
    images in terms of pixel randomness, as quantified by image entropy. 
    \item We propose \textbf{M}ulti-granularity \textbf{L}ocal \textbf{E}ntropy \textbf{P}atterns (\textbf{MLEP}), a set of feature maps with entropy computed from shuffled small patches across multiple resampling scales. MLEP effectively disrupts image semantics to mitigate content bias, while capturing pixel relationships across both spatial and scale dimensions. Using MLEP as input, a standard CNN classifier can be trained for robust and generalized AIGI detection.
    \item{Extensive quantitative and qualitative analyses validate the effectiveness of the MLEP design, showing significant improvements over state-of-the-art methods across multiple AI-generated image datasets.}
\end{itemize}

\section{Related Work}
Detecting AI-generated images typically relies on three types of cues: spatial-domain cues, frequency-domain cues, and representations from pretrained models. 

\paragraph{Spatial-domain Detection}
Spatial-domain methods typically rely on handcrafted spatial features, local patterns, or pixel statistics to distinguish between real and generated images. 
The representative methods include generalized feature extraction from CNN-based model~\cite{wang2020cnn} and inter-pixel correlation between rich and poor texture regions~\cite{zhong2024patchcraft}. 
Recently, Tan et al.~\cite{tan2024rethinking} observed that upsampling operations are prevalent in image generation models and proposed utilizing neighboring pixel relationships (NPR), computed through local pixel differences, as a simple yet effective cue for AIGI detection.
Zheng et al.~\cite{zheng2024breaking} discovered that image semantic information negatively impacts detection performance and proposed a simple linear classifier that utilizes image patch shuffling to disrupt the original semantic artifacts.

\paragraph{Frequency-domain Detection}
To tackle the subtlety of spatial artifacts in AI-generated images, frequency-domain methods analyze image frequency components, enabling more effective real vs. fake differentiation. 
The study in~\cite{frank2020leveraging} found that 
GAN-generated images exhibit generalized artifacts in discrete cosine transform (DCT) spectrum, 
which can be readily identified. 
Qian et al.~\cite{qian2020thinking} proposed a face forgery detection network based on frequency-aware decomposed image components and local frequency statistics.
Luo et al.~\cite{luo2021generalizing} proposed a feature representation based on high-frequency noise at multiple scales and enhanced detection performance by integrating it with an attention module.
Liu et al.~\cite{liu2022detecting} utilized noise patterns in the frequency domain as feature representations for detecting AI-generated images. 
Tan et al.~\cite{tan2024frequency} proposed FreqNet toward detection generalizability, which focuses on high-frequency components of images, exploiting high-frequency representation across spatial and channel dimension.

\paragraph{Detection leveraging Pretrained Models}
This group of methods aims to derive generalized features for AIGI detection by leveraging the knowledge learned by large models pretrained on extensive datasets. 
Wang et al.~\cite{wang2023dire} proposed an artifact representation named DIffusion Reconstruction Error (DIRE), which obtains the difference between the input image and its reconstructed object through a pretrained diffusion model.
Chen et al.~\cite{chen2024drct} proposed utilizing pretrained diffusion models to generate high-quality synthesized images, serving as challenging samples to enhance the detector’s performance. 
Ojha et al.~\cite{ojha2023towards} utilized representations from a fixed pretrained CLIP model as generalized features for detection. 
Recently, studies such as~\cite{khan2024clipping,liu2024forgery} leveraged textual information from vision-language models to further enhance detection performance.

\section{The Approach} \label{sec:method}
\begin{figure*}[t]
    \centering
    \includegraphics[width=\textwidth]{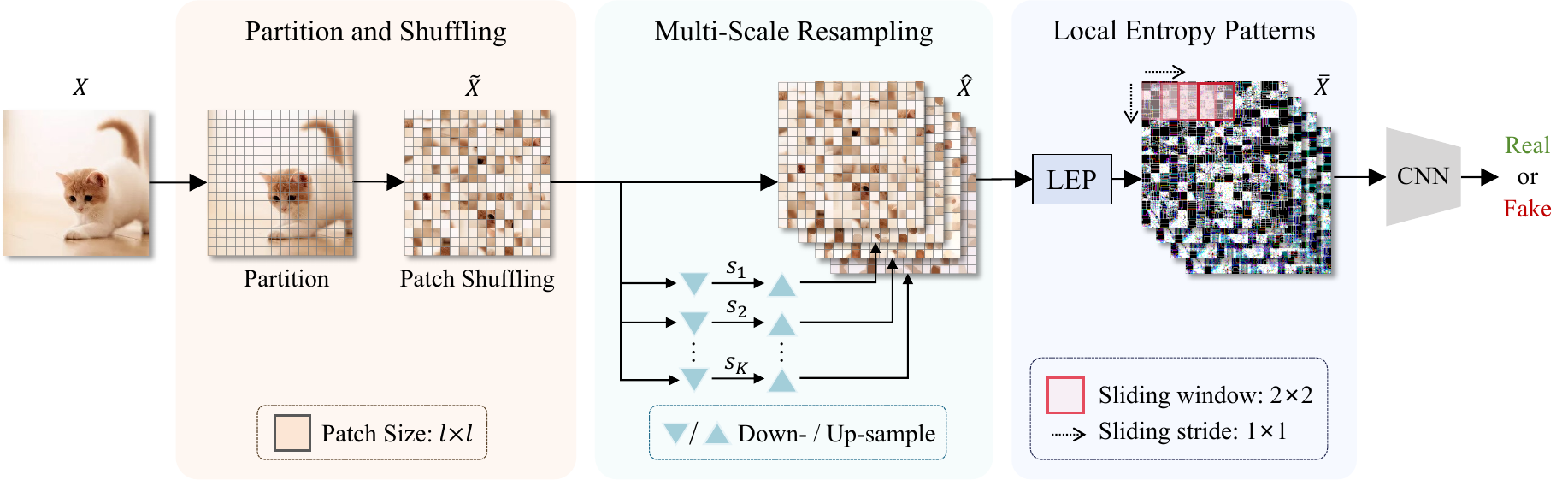}
    \caption{The MLEP framework, composed of three steps: Patch Shuffling, Multi-Scale Resampling, and Local Entropy Patterns computation.}
    \label{overview}
\end{figure*}

The proposed approach leverages entropy-based feature extraction to analyze local pixel randomness in a multi-granularity, semantic-agnostic manner. It begins by dividing the image into small patches and applying random shuffling to reduce semantic bias. A multi-scale pyramid is then constructed by downsampling and upsampling the scrambled image, introducing resampling artifacts. Local entropy is computed using a $2\times2$ sliding window across the entire image, capturing complexity across intra-block, inter-block, and inter-scale levels. The resulting multi-granularity local entropy patterns (MLEP) are used as input to a standard CNN classifier for distinguishing AI-generated from real images. An overview of the method is shown in Fig.~\ref{overview}, with key components detailed below.

\subsection{Semantic Suppression via Patch Shuffling}
Inspired by previous work~\cite{zhong2024patchcraft,zheng2024breaking} that mitigates semantic bias via patch-based processing, we adopt finer patch shuffling to further disrupt image content. Given an input image $X \in \mathbb{R}^{H\times W \times C}$, we divide it into uniform $l \times l$ patches:
\begin{equation}
\label{deqn_ex1}
    X = \{ X_{i,j} \in \mathbb{R}^{l\times l \times C} \}_{1 \leq i \leq \frac{H}{l}, 1 \leq j \leq \frac{W}{l}},
\end{equation}
where $l$ is a small integer (typically $<8$), and $H, W$ are assumed divisible by $l$. 
The patches are then randomly permuted, 
resulting in a visually scrambled image denoted as $\tilde{X}$:
\begin{equation}
\label{eq:x_shuffle}
    \tilde{X} = \{ \tilde{X}_{\pi(i,j)}=X_{i,j} \}_{1 \leq i \leq \frac{H}{l}, 1 \leq j \leq \frac{W}{l}},
\end{equation}
where $\pi$ is a bijection defining the patch permutation. Note that partitioning and shuffling are applied independently to each color channel.

\begin{figure}
    \centering
    \subfloat[LEP of a single $2\times 2$ window.]{
        \includegraphics[width=0.4\columnwidth]{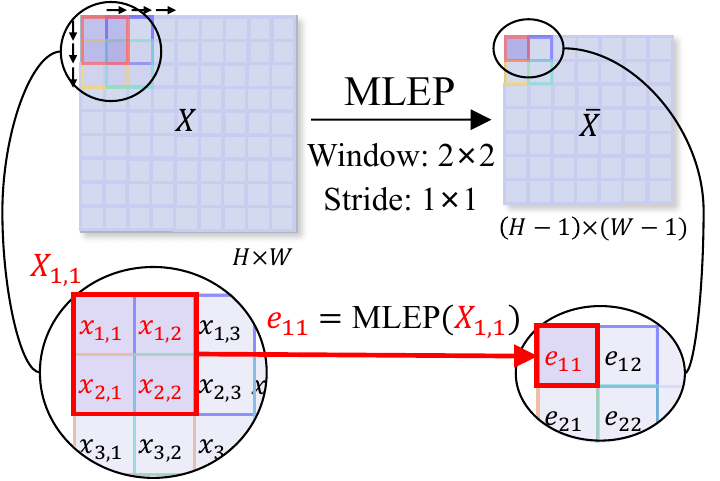}
    }
    \hfil
    \subfloat[Intra- and Inter-block LEP.]{
	\includegraphics[width=0.38\columnwidth]{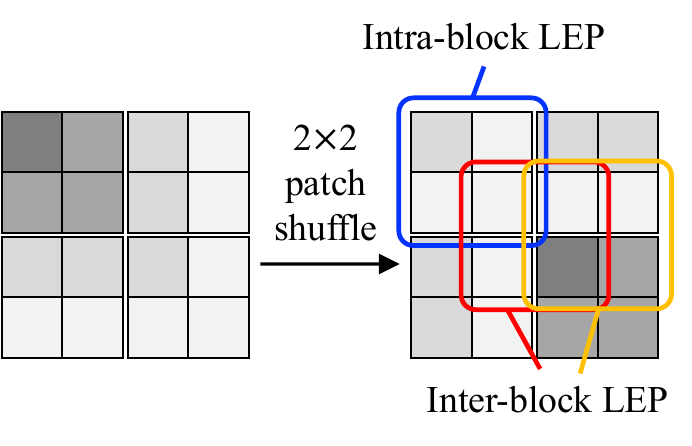}
    }
    \caption{Illustration of LEP computation in a single patch (a) and its intra-block and inter-block views (b) when using a $2\times 2$ sliding window with stride of 1 across shuffled image patches.}
    \label{fig:entropy_illustration}
\end{figure}

\subsection{Multi-Scale Resampling} 
Inspired by~\cite{tan2024rethinking} showing that generative models often use upsampling to produce high-resolution outputs, we propose detecting generation artifacts via multi-scale analysis. We hypothesize that resampling generated images reveals distinctive patterns useful for detection. To this end, we first construct a multi-scale pyramid by resampling the scrambled image $\tilde{X}$ with scale factors $\mathbb{S} = \{s_1, s_2, \dots, s_K\}$, with each scale $s_k \in (0,1]$ applied using an interpolation function $\mathrm{Down}(\cdot, s_k)$:
\begin{equation} \label{eq:downsample}
    \tilde{X}_{\vee}^{(k)} = \mathrm{Down}(\tilde{X}, s_k), \quad 
    \tilde{X}_{\vee}^{(k)} \in \mathbb{R}^{\lfloor s_k \cdot H\rfloor \times \lfloor s_k \cdot W\rfloor \times C},
\end{equation}
which are then upsampled back to its original shape using an interpolation function 
$\mathrm{Up}(\cdot, H, W)$:
\begin{equation} \label{eq:upsample}
    \tilde{X}_{\wedge}^{(k)} = \mathrm{Up}(\tilde{X}_{\vee}^{(k)}, H, W), \quad 
    \tilde{X}_{\wedge}^{(k)} \in \mathbb{R}^{H \times W \times C}.
\end{equation} 
The resulting multi-scale resampling image $\hat{X}$ is created by concatenating all the upsampled images along the channel dimension: 
\begin{equation} \label{eq:pyramid}
    \hat{X} = \mathrm{Concat}(\tilde{X}_{\wedge}^{(1)}, \tilde{X}_{\wedge}^{(2)}, \dots, \tilde{X}_{\wedge}^{(K)}), 
    \quad \hat{X} \in \mathbb{R}^{H \times W \times (C\cdot K)}.
\end{equation}

\subsection{Multi-granularity Local Entropy Patterns}
The core of our approach is the design of Local Entropy Patterns (LEP), which quantify textural randomness using a $2\times2$ sliding window over pixel sets $\hat{X}_{i,j} = \{x_{m,n}\}_{m \in \{i,i+1\}, n \in \{j,j+1\}}$, based on Shannon's definition of information entropy~\cite{shannon1948mathematical}: 
\begin{equation}
    \mathrm{LEP}\left(\hat{X}_{i,j}\right) = - \sum_{m,n} p(x_{m,n}) \cdot \log_2 p(x_{m,n}), 
\end{equation}
where $p(x_{m,n})$ represents the probability of occurrence of the pixel value $x_{m,n}$ within that specific patch $\hat{X}_{i,j}$. 
By restricting the sliding window to $2\times2$ (four pixels), entropy values are confined to five discrete levels: $\mathbb{V} = \{0, 0.8, 1.0, 1.5, 2\}$. The proof and an efficient computation algorithm for LEP on a $2\times2$ window are provided in the supplementary material. 
With a stride of 1, the $2 \times 2$ sliding window introduces overlap in LEP computation. Due to patch shuffling, this captures both \textit{intra-patch} and \textit{inter-patch} entropy—reflecting local randomness within and across original image regions—as illustrated in Fig.~\ref{fig:entropy_illustration}. Applied across multiple scales, LEP further captures \textit{inter-scale} entropy, forming the basis of the final Multi-granularity Local Entropy Patterns (MLEP).


\subsection{AIGI Detection based on MLEP}
Given the computed MLEP feature maps denoted as $\bar{X} \in \mathbb{V}^{(H-1) \times (W-1) \times (C\cdot K)},$
a representative CNN-based classifier can be trained 
to differentiate between photographic and AI-generated images. 
Denoting the classifier as $f$, the training objective 
is defined using the binary cross-entropy loss:
\begin{equation}
    \mathcal{L}_{BCE} = -\frac{1}{N}\sum_{i=1}^N
    \left[y_i\log(f(\bar{X}_i)) + (1-y_i)\log(1 - f(\bar{X}_i))\right],
\end{equation}
where $y_i$ represents the true labels, 
$f(\bar{X}_i)$ the predictions, 
and $N$ the number of training samples.

\section{Experiments} \label{sec:experiments}

\begin{table}
    \caption{Detection performance in terms of  Acc.(\%) and A.P.(\%) on the GAN-based datasets. }
    \label{evaluation table on GAN-set}
    \centering
    \resizebox{\textwidth}{!}{ 
    \begin{tabular}{lrrrrrrrrrrrrrrrr}
        \toprule
        \multirow{2}{*}{Method} & \multicolumn{2}{c}{ProGAN} & \multicolumn{2}{c}{StyleGAN} & \multicolumn{2}{c}{StyleGAN2} & \multicolumn{2}{c}{BigGAN} & \multicolumn{2}{c}{CycleGAN} & \multicolumn{2}{c}{StarGAN} & \multicolumn{2}{c}{GauGAN} & \multicolumn{2}{c}{AttGAN} \\
                 & Acc. & A.P.  & Acc. & A.P.  & Acc. & A.P.  & Acc. & A.P. & Acc. & A.P. & Acc.  & A.P.  & Acc. & A.P.  & Acc. & A.P. \\
        \midrule
        CNNDet \cite{wang2020cnn}  & 91.4 & 99.4  & 63.8 & 91.4  & 76.4 & 97.5  & 52.9 & 73.3 & 72.7 & 88.6 & 63.8  & 90.8  & 63.9 & 92.2  & 51.1  & 83.7 \\
        F3Net \cite{qian2020thinking}   & 99.4 & 100.0 & 92.6 & 99.7  & 88.0 & 99.8  & 65.3 & 69.9 & 76.4 & 84.3 & 100.0 & 100.0 & 58.1 & 56.7  & 85.2  & 94.8 \\
        LGrad \cite{tan2023learning}   & 99.0 & 100.0 & 94.8 & 99.9  & 96.0 & 99.9  & 82.9 & 90.7 & 85.3 & 94.0 & 99.6  & 100.0 & 72.4 & 79.3  & 68.6  & 93.8  \\
        Ojha \cite{ojha2023towards}    & 99.7 & 100.0 & 89.0 & 98.7  & 83.9 & 98.4  & 90.5 & 99.1 & 87.9 & 99.8 & 91.4  & 100.0 & 89.9 & 100.0 & 78.5  & 91.3  \\
        Zheng \cite{zheng2024breaking} & 99.7 & 100.0 & 90.7 & 95.3 & 97.6 & 99.7 & 67.0 & 67.6 & 85.2 & 92.6 & 98.7 & 100.0 & 57.1 & 56.8 & 79.4 & 87.7 \\
        CLIPping \cite{khan2024clipping} & 99.8 & 100.0 & 94.3 & 99.4  & 83.5 & 98.7  & 93.8 & 99.4 & 95.4 & 99.9 & 99.1  & 100.0 & 93.4 & 99.9  & 91.3  & 97.4  \\
        NPR \cite{tan2024rethinking}     & 99.8 & 100.0 & 96.3 & 99.8  & 97.3 & 100.0 & 87.5 & 94.5 & 95.0 & 99.5 & 99.7  & 100.0 & 86.6 & 88.8  & 83.0  & 96.2  \\
        FreqNet \cite{tan2024frequency} & 99.6 & 100.0 & 90.2 & 99.7 & 87.9 & 99.5 & 90.5 & 96.0 & 95.8 & 99.6 & 85.6 & 99.8 & 93.4 & 98.6 & 89.8 & 98.8 \\
        FatFormer \cite{liu2024forgery} & \textbf{99.9} & 100.0 & 97.1 & 99.8 & 98.8 & 99.9 & \textbf{99.5} & \textbf{100.0} & \textbf{99.4} & \textbf{100.0} & 99.8 & 100.0 & \textbf{99.4} & \textbf{100.0} & 99.3 & 100.0 \\
        \rowcolor{gray! 30}
        Ours     & 99.6 & \textbf{100.0} & \textbf{99.6} & \textbf{100.0} & \textbf{99.9} & \textbf{100.0} & 87.1 & 93.6 & 98.3 & 99.3 & \textbf{100.0} & \textbf{100.0} & 82.0 & 87.9 & \textbf{100.0} & \textbf{100.0} \\
        \bottomrule
    \end{tabular}
    }

    \resizebox{\textwidth}{!}{ 
    \begin{tabular}{lrrrrrrrrrrrrrrrr}
        \toprule
        \multirow{2}{*}{Method} & \multicolumn{2}{c}{BEGAN} & \multicolumn{2}{c}{CramerGAN} & \multicolumn{2}{c}{InfoMaxGAN} & \multicolumn{2}{c}{MMDGAN} & \multicolumn{2}{c}{RelGAN} & \multicolumn{2}{c}{S3GAN} & \multicolumn{2}{c}{SNGAN} & \multicolumn{2}{c}{STGAN}  \\
                 & Acc.  & A.P.  & Acc.  & A.P.  & Acc. & A.P.  & Acc. & A.P.  & Acc.  & A.P.  & Acc. & A.P. & Acc. & A.P. & Acc. & A.P.  \\
         \midrule
        CNNDet \cite{wang2020cnn}  & 50.2  & 44.9  & 81.5  & 97.5  & 71.1 & 94.7  & 72.9 & 94.4  & 53.3  & 82.1  & 55.2 & 66.1 & 62.7 & 90.4 & 63.0 & 92.7  \\
        F3Net \cite{qian2020thinking}   & 87.1  & 97.5  & 89.5  & 99.8  & 67.1 & 83.1  & 73.7 & 99.6  & 98.8  & 100.0 & 65.4 & 70.0 & 51.6 & 93.6 & 60.3 & 99.9  \\
        LGrad \cite{tan2023learning}   & 69.9  & 89.2  & 50.3  & 54.0  & 71.1 & 82.0  & 57.5 & 67.3  & 89.1  & 99.1  & 78.5 & 86.0 & 78.0 & 87.4 & 54.8 & 68.0  \\
        Ojha \cite{ojha2023towards}    & 72.0  & 98.9  & 77.6  & 99.8  & 77.6 & 98.9  & 77.6 & 99.7  & 78.2  & 98.7  & 85.2 & 98.1 & 77.6 & 98.7 & 74.2 & 97.8  \\
        Zheng \cite{zheng2024breaking} & 67.4 & 98.0 & 74.2 & 93.8 & 71.0 & 93.1 & 68.4 & 89.4 & 98.4 & 99.9 & 70.8 & 69.9 & 72.4 & 94.0 & 92.3 & 100.0 \\
        CLIPping \cite{khan2024clipping} & \textbf{100.0} & 100.0 & \textbf{100.0} & 100.0 & 94.7 & 99.7  & 94.8 & 99.9  & 92.2  & 98.3  & 88.4 & 97.7 & 94.4 & 99.5 & 87.2 & 96.4  \\
        NPR \cite{tan2024rethinking}     & 99.0  & 99.8  & 98.7  & 99.0  & 94.5 & 98.3  & 98.6 & 99.0  & 99.6  & 100.0 & 79.0 & 80.0 & 88.8 & 97.4 & 98.0 & 100.0 \\
        FreqNet \cite{tan2024frequency} & 98.8 & 100.0 & 95.1 & 98.2 & 94.5 & 97.3 & 95.1 & 98.2 & 100.0 & 100.0 & 88.4 & 94.3 & 85.3 & 90.5 & 98.8 & 100.0 \\
        FatFormer \cite{liu2024forgery} & 99.9 & 100.0 & 98.4 & \textbf{100.0} & \textbf{98.4} & \textbf{100.0} & 98.4 & \textbf{100.0} & 99.5 & 100.0 & \textbf{99.0} & \textbf{100.0} & \textbf{98.3} & \textbf{99.9} & 98.8 & 99.8 \\
        \rowcolor{gray! 30} 
        Ours     & 99.4 & \textbf{100.0} & 98.5 & 99.8 & 98.0 & 99.8 & \textbf{98.9} & 99.8 & \textbf{100.0} & \textbf{100.0} & 83.4 & 91.7 & 97.6 & 99.7 & \textbf{99.9} & \textbf{100.0} \\
        \bottomrule
        \end{tabular}
        }
\end{table}

\begin{table}
    \caption{Detection performance in terms of  Acc.(\%) and A.P.(\%) on the Diffusion-based datasets. }
    \label{evaluation table on Diffusion-set}
    \centering
    \resizebox{\textwidth}{!}{ 
        \begin{tabular}{lrrrrrrrrrrrrrrrr}
        \toprule
        \multirow{2}{*}{Method} & \multicolumn{2}{c}{ADM} & \multicolumn{2}{c}{DDPM} & \multicolumn{2}{c}{IDDPM} & \multicolumn{2}{c}{LDM} & \multicolumn{2}{c}{PNDM} & \multicolumn{2}{c}{VQ-Diffusion} & \multicolumn{2}{c}{SDv1} & \multicolumn{2}{c}{SDv2} \\
                 & Acc. & A.P. & Acc. & A.P.  & Acc. & A.P.  & Acc.  & A.P.  & Acc. & A.P.  & Acc.  & A.P.  & Acc. & A.P. & Acc. & A.P.  \\
         \midrule
        CNNDet \cite{wang2020cnn}  & 53.9 & 71.8 & 62.7 & 76.6  & 50.2 & 82.7  & 50.4  & 78.7  & 50.8 & 90.3  & 50.0  & 71.0  & 38.0 & 76.7 & 52.0 & 90.3  \\
        F3Net \cite{qian2020thinking}   & 80.9 & 96.9 & 84.7 & 99.4  & 74.7 & 98.9  & 100.0 & 100.0 & 72.8 & 99.5  & 100.0 & 100.0 & 73.4 & 97.2 & 99.8 & 100.0 \\
        LGrad \cite{tan2023learning}   & 86.4 & 97.5 & 99.9 & 100.0 & 66.1 & 92.8  & 99.7  & 100.0 & 69.5 & 98.5  & 96.2  & 100.0 & 90.4 & 99.4 & 97.1 & 100.0 \\
        Ojha \cite{ojha2023towards}    & 78.4 & 92.1 & 72.9 & 78.8  & 75.0 & 92.8  & 82.2  & 97.1  & 75.3 & 92.5  & 83.5  & 97.7  & 56.4 & 90.4 & 71.5 & 92.4  \\
        Zheng \cite{zheng2024breaking} & 72.1 & 78.9 & 78.9 & 80.5 & 49.9 & 52.0 & 99.7 & 100.0 & 90.4 & 96.9 & 99.6 & 100.0 & 94.0 & 99.7 & 87.9 & 96.4 \\
        CLIPping \cite{khan2024clipping} & 78.9 & 93.8 & 80.3 & 85.7  & 82.4 & 94.4  & 90.2  & 97.6  & 81.7 & 93.7  & 96.3  & 99.3  & 58.0 & 93.1 & 82.6 & 94.9  \\
        NPR \cite{tan2024rethinking}     & 88.6 & 98.9 & 99.8 & 100.0 & 91.8 & 99.8  & \textbf{100.0} & 100.0 & 91.2 & 100.0 & 100.0 & 100.0 & 97.4 & 99.8 & 93.8 & 100.0 \\
        FreqNet \cite{tan2024frequency} & 67.2 & 91.3 & 91.5 & 99.8 & 59.0 & 97.3 & 98.9 & 100.0 & 85.2 & 99.8 & 100.0 & 100.0 & 63.9 & 98.1 & 81.8 & 98.4 \\
        FatFormer \cite{liu2024forgery} & 70.8 & 93.4 & 67.2 & 72.5 & 69.3 & 94.3 & 97.3 & 100.0 & 99.3 & 100.0 & 100.0 & 100.0 & 61.7 & 96.8 & 84.4 & 98.2 \\
        \rowcolor{gray! 30}
        Ours     & \textbf{97.0} & \textbf{99.8} & \textbf{100.0} & \textbf{100.0} & \textbf{100.0} & \textbf{100.0} & 99.8 & \textbf{100.0} & \textbf{100.0} & \textbf{100.0} & \textbf{100.0} & \textbf{100.0} & \textbf{98.5} & \textbf{99.9} & \textbf{100.0} & \textbf{100.0} \\
        \bottomrule
        \end{tabular}
    }

    \centering
    \resizebox{\textwidth}{!}{ 
        \begin{tabular}{lrrrrrrrrrrrrrrrr}
        \toprule
        \multirow{2}{*}{Method} & \multicolumn{2}{c}{DALL·E mini} & \multicolumn{2}{c}{Glide-100-10} & \multicolumn{2}{c}{Glide-100-27} & \multicolumn{2}{c}{Glide-50-27} & \multicolumn{2}{c}{LDM-200} & \multicolumn{2}{c}{LDM-200-cfg} & \multicolumn{2}{c}{Midjourney} & \multicolumn{2}{c}{DALL·E 2} \\
                 & Acc. & A.P. & Acc. & A.P.  & Acc. & A.P.  & Acc. & A.P.  & Acc. & A.P.  & Acc. & A.P. & Acc. & A.P.  & Acc. & A.P. \\
        \midrule
        CNNDet \cite{wang2020cnn}  & 51.8 & 61.3 & 53.3 & 72.9  & 53.0 & 71.3  & 54.2 & 76.0  & 52.0 & 64.5  & 51.6 & 63.1  & 48.6 & 38.5 & 49.3 & 44.7 \\
        F3Net \cite{qian2020thinking}   & 71.6 & 79.9 & 88.3 & 95.4  & 87.0 & 94.5  & 88.5 & 95.4  & 73.4 & 83.3  & 80.7 & 89.1  & 73.2 & 80.4 & 79.6 & 87.3 \\
        LGrad \cite{tan2023learning}   & 88.5 & 97.3 & 89.4 & 94.9  & 87.4 & 93.2  & 90.7 & 95.1  & 94.2 & 99.1  & 95.9 & 99.2  & 68.3 & 76.0 & 75.1 & 80.9 \\
        Ojha \cite{ojha2023towards}    & 89.5 & 96.8 & 90.1 & 97.0  & 90.7 & 97.2  & 91.1 & 97.4  & 90.2 & 97.1  & 77.3 & 88.6  & 50.0 & 49.8 & 66.3 & 74.6 \\
        Zheng \cite{zheng2024breaking} & 67.9 & 72.2 & 79.4 & 87.8 & 76.8 & 84.5 & 78.2 & 85.9 & 81.3 & 90.1 & 84.0 & 91.7 & 73.2 & 78.5 & 81.4 & 89.2 \\
        CLIPping \cite{khan2024clipping} & 91.1 & 98.6 & 92.0 & 98.6  & 91.2 & 98.8  & 94.3 & 99.3  & 92.8 & 98.9  & 77.4 & 94.3  & 51.1 & 50.7 & 62.6 & 72.3 \\
        NPR \cite{tan2024rethinking}     & 94.5 & 99.5 & 98.2 & 99.8  & 97.8 & 99.7  & 98.2 & 99.8  & 99.1 & 99.9  & 99.0 & 99.8  & 77.4 & 81.9 & 80.7 & 83.0 \\
        FreqNet \cite{tan2024frequency} & 97.4 & 99.8 & 88.1 & 96.4 & 84.5 & 96.1 & 86.7 & 96.3 & 97.5 & 99.9 & 97.4 & 99.9 & 55.5 & 65.3 & 52.9 & 61.8 \\
        FatFormer \cite{liu2024forgery} & \textbf{98.8} & 99.8 & 94.2 & 99.2 & 94.4 & 99.1 & 94.7 & 99.4 & 98.6 & 99.8 & 94.9 & 99.1 & 62.8 & 85.4 & 68.8 & 93.2 \\
        \rowcolor{gray! 30}
        Ours     & 95.7 & \textbf{99.9} & \textbf{99.9} & \textbf{100.0} & \textbf{100.0} & \textbf{100.0} & \textbf{99.8} & \textbf{100.0} & \textbf{99.9} & \textbf{100.0} & \textbf{99.8} & \textbf{100.0} & \textbf{87.5} & \textbf{97.1} & \textbf{87.3} & \textbf{97.4} \\
        \bottomrule
        \end{tabular}
    }
\end{table}

\subsection{Experimental Settings}
\paragraph{Datasets}
To evaluate generality, we adopt the cross-dataset setup from~\cite{tan2024rethinking}. 
We use the ForenSynths~\cite{wang2020cnn} dataset for training, 
which contains 20 content categories, 
each with 18,000 synthetic images generated by ProGAN~\cite{karras2018progressive} 
and an equal number of real images from LSUN~\cite{yu2015lsun}. 
Following the setup of~\cite{tan2024rethinking}, 
we restrict the training to only four categories: 
\textit{cars}, \textit{cats}, \textit{chairs}, and \textit{horses}, 
thus posing a more challenging cross-scene detection problem. 
Following~\cite{wang2020cnn,wang2023dire,tan2024rethinking,zheng2024breaking}, 
we evaluate the proposed method using synthesized images from 
32 different models, involving 16 GAN-based models 
and 16 Diffusion-based models (including their variants). 
Details about the synthesized image sources are provided below:

The \textbf{GAN-Set} includes 
ProGAN~\cite{karras2018progressive}, 
StyleGAN~\cite{karras2019style}, 
StyleGAN2~\cite{karras2020analyzing}, 
BigGAN~\cite{brock2018large}, 
CycleGAN~\cite{zhu2017unpaired}, 
StarGAN~\cite{choi2018stargan}, 
GauGAN~\cite{park2019semantic}, 
AttGAN~\cite{he2019attgan}, 
BEGAN~\cite{berthelot2017began}, 
CramerGAN~\cite{bellemare2017cramer}, 
InfoMaxGAN~\cite{lee2021infomax}, 
MMDGAN~\cite{li2017mmd}, 
RelGAN~\cite{nie2018relgan}, 
S3GAN~\cite{luvcic2019high}, 
SNGAN~\cite{miyato2018spectral}, 
and STGAN~\cite{liu2019stgan}, 
with the former seven obtained from 
the dataset ForenSynths~\cite{wang2020cnn} and the latter nine from 
the dataset GANGen-Detection~\cite{tan2024GANGen-Detection}.

The \textbf{Diffusion-Set} contains  
DDPM~\cite{ho2020denoising}, 
IDDPM~\cite{nichol2021improved}, 
ADM~\cite{dhariwal2021diffusion}, 
LDM~\cite{rombach2022high}, 
PNDM~\cite{liu2022pseudo}, 
VQ-Diffusion~\cite{gu2022vector}, 
Stable Diffusion (SD) v1/v2~\cite{rombach2022high}, 
DALL·E mini~\cite{ramesh2021zero}, 
three Glide~\cite{nichol2022glide} variants\footnote{Glide with 100 steps in the first stage and 10 steps in the second stage (aka Glide-100-10), Glide-100-27, and Glide-50-27}, 
and two LDM~\cite{rombach2022high} variants\footnote{LDM with 200 steps (LDM-200), and LDM with 200 steps with classifier-free diffusion guidance (LDM-200-CFG)}. 
Of these models, the first eight are sourced from the DiffusionForensics dataset~\cite{wang2023dire}, while the remainder are from the UniversalFakeDetect dataset~\cite{ojha2023towards}. Furthermore, we include images from two commercial models, 
Midjourney and DALL·E 2, sourced from the social platform 
Discord\footnote{\url{https://discord.com/}} as provided by~\cite{tan2024rethinking}.

Each above subset comprises an equal number of real samples paired with the corresponding generative counterparts. All test images were obtained according to the 
instructions provided by~\cite{tan2024rethinking}.

\paragraph{Implementation Details}
All images were resized to $224 \times 224$, with random cropping for training and center cropping for testing. To evaluate the effects of patch size ($l$) and resampling scales ($\mathbb{S}$), we tested $l \in \{2, 4, 8\}$ and $\mathbb{S} = \{1, 1/2, 1/4, 1/8\}$. Bilinear interpolation was used for multi-scale resampling, with nearest and bicubic methods also evaluated. ResNet-50~\cite{he2016resnet} served as the primary backbone, with additional tests on ResNet-18, ResNet-34, and ResNet-101. The model was trained using Adam optimizer (learning rate 0.002, batch size 64). All experiments ran on a server with two NVIDIA RTX A5000 GPUs.

\paragraph{Baseline Methods}
We compare against representative baselines, including CNNDet~\cite{wang2020cnn}, F3Net~\cite{qian2020thinking}, LGrad~\cite{tan2023learning}, UnivFD~\cite{ojha2023towards}, CLIPping~\cite{khan2024clipping}, NPR~\cite{tan2024rethinking}, Zheng~\cite{zheng2024breaking}, FreqNet~\cite{tan2024frequency}, and FatFormer~\cite{liu2024forgery}. Accuracy (Acc.) and average precision (A.P.) are used as metrics. CLIPping~\cite{khan2024clipping}, Zheng~\cite{zheng2024breaking}, FreqNet~\cite{tan2024frequency}, and FatFormer~\cite{liu2024forgery} were re-evaluated under the same protocol, while results for other baselines were taken from~\cite{tan2024rethinking} under consistent settings.

\subsection{Evaluation Results}
\subsubsection{Evaluation of Detection Generalization Capability}

\begin{wraptable}[17]{r}{0.55\textwidth}
    \tabcolsep = 3pt
    \centering
    \caption{Mean Acc. and A.P. over 16 GAN-based, 16 Diffusion-based, and all 32 datasets.}
        \begin{tabular}{lrrrrrr}
        \toprule
        \multirow{2}{*}{Method} & \multicolumn{2}{c}{GAN-Set} & \multicolumn{2}{c}{Diff.-Set} & \multicolumn{2}{c}{Mean}  \\
                 & Acc. & A.P. & Acc. & A.P. & Acc. & A.P. \\
         \midrule
        CNNDet \cite{wang2020cnn}  & 65.4 & 86.2 & 51.4 & 70.7 & 58.4 & 78.4 \\
        F3Net \cite{qian2020thinking}   & 78.7 & 90.6 & 83.0 & 93.6 & 80.8 & 92.1 \\
        LGrad \cite{tan2023learning}   & 78.0 & 86.9 & 87.2 & 95.2 & 82.6 & 91.1 \\
        Ojha  \cite{ojha2023towards}   & 83.2 & 98.6 & 77.5 & 89.5 & 80.4 & 94.1 \\
        Zheng \cite{zheng2024breaking}   & 80.6 & 89.9 & 80.9 & 86.5 & 80.8 & 88.2 \\
        CLIPping \cite{khan2024clipping} & 93.9 & 99.1 & 81.4 & 91.5 & 87.7 & 95.3 \\
        NPR \cite{tan2024rethinking}     & 93.8 & 97.0 & 94.2 & 97.6 & 94.0 & 97.3 \\
        FreqNet \cite{tan2024frequency} & 93.1 & 98.2 & 81.7 & 93.8 & 87.4 & 96.0 \\
        FatFormer \cite{liu2024forgery}  & \textbf{99.0} & \textbf{100.0} & 84.8 & 95.6 & 91.9 & 97.8 \\
        \rowcolor{gray! 30}
        Ours     & 96.4 & 98.2 & \textbf{97.8} & \textbf{99.6} & \textbf{97.1} & \textbf{98.9} \\
        \bottomrule
        \end{tabular}
    \label{mean evaluation table}
\end{wraptable}

We evaluated the generalization performance of our AIGI detection method across datasets. Accuracy (Acc.) and average precision (A.P.) compared to state-of-the-art GAN- and Diffusion-based methods are reported in Tables~\ref{evaluation table on GAN-set} and~\ref{evaluation table on Diffusion-set}, using patch size $l=2$, scales $\mathbb{S} = \{1, 1/2, 1/4\}$, and a ResNet-50 backbone. MLEP consistently achieves top performance across most datasets. Remarkably, it generalizes well to diffusion-generated images, despite being trained solely on GAN-based data (ProGAN~\cite{karras2018progressive}). Even on datasets with entirely different content (e.g., face-centric sets like StarGAN, InfoMaxGAN, and AttGAN), MLEP maintains strong performance, underscoring its cross-scene robustness. Table~\ref{mean evaluation table} further shows that MLEP outperforms NPR~\cite{tan2024rethinking}, with average gains of 3.1\% in Acc. and 1.6\% in A.P., despite NPR's already strong results.

\subsubsection{Ablation Study}
We next conducted a series of ablation studies to evaluate the effectiveness of key components and hyperparameters in the proposed approach.

\begin{wraptable}[11]{r}{0.55\textwidth}
    \tabcolsep = 3.5pt
    \centering
    \caption{Ablation study on the impact of key components, where PS represents patch shuffling and MR denotes multi-scale resampling.}
    \label{ablation study table}
        \begin{tabular}{cccrrrrrr}
        \toprule
        \multirow{2}{*}{LEP} & \multirow{2}{*}{PS} & \multirow{2}{*}{MR} & \multicolumn{2}{c}{GAN-set} & \multicolumn{2}{c}{Diff.-set} & \multicolumn{2}{c}{Mean} \\
        &&& Acc. & A.P. & Acc. & A.P. & Acc. & A.P. \\
         \midrule
        \checkmark &            &            & 93.6 & 94.1 & 94.9 & 95.7 & 94.3 & 94.9 \\
        \checkmark &            & \checkmark & 93.4 & 94.1 & 95.8 & 96.9 & 94.6 & 95.5 \\
        \checkmark & \checkmark &            & 95.7 & 98.2 & 97.5 & 99.6 & 96.6 & 98.9 \\
        \rowcolor{gray! 30}
        \checkmark & \checkmark & \checkmark & \textbf{96.4} & \textbf{98.2} & \textbf{97.8} & \textbf{99.6} & \textbf{97.1} & \textbf{98.9} \\
        \bottomrule
        \end{tabular}
\end{wraptable}

\paragraph{Effectiveness of patch shuffling and multi-scale resampling}
We first assessed the impact of two key components: patch shuffling and multi-scale resampling. Ablation results in Table~\ref{ablation study table} show that removing either component noticeably reduces performance, with patch shuffling contributing more. Even without both, LEP alone achieves over 94.3\% accuracy, higher than NPR (94.0\%)~\cite{tan2024rethinking}, highlighting the effectiveness of entropy-based features.

\begin{wraptable}[16]{r}{0.68\textwidth}
\tabcolsep=0.3em
    \caption{Impact of patch size $l$ and resampling scaling factors $\mathbb{S}$.}
    \label{evaluation table on different hyperparameters}
    \centering
        \begin{tabular}{cl|cccccc}
        \toprule
        \multirow{2}{*}{Patch size} & \multirow{2}{*}{Scale factors $\mathbb{S}$} & \multicolumn{2}{c}{GAN-set} & \multicolumn{2}{c}{Diff.-set} & \multicolumn{2}{c}{Mean}  \\
         & & Acc. & A.P. & Acc. & A.P. & Acc. & A.P. \\
         \midrule
                & $\{1, 1/2\}$             & 95.8 & 97.8 & 97.5 & 99.6 & 96.6 & 98.7 \\
         \rowcolor{gray! 30}
          $l=2$ & $\{1, 1/2, 1/4\}$       & \textbf{96.4} & \textbf{98.2} & \textbf{97.8} & \textbf{99.6} & \textbf{97.1} & \textbf{98.9} \\
         & $\{1, 1/2, 1/4, 1/8\}$ & 91.7 & 97.9 & 95.3 & 99.5 & 93.5 & 98.7 \\
        \midrule
               & $\{1, 1/2\}$             & 94.5 & 96.6 & 95.5 & 98.8 & 95.0 & 97.7 \\
         $l=4$ & $\{1, 1/2, 1/4\}$       & 94.5 & 96.8 & 96.6 & 99.1 & 95.5 & 97.9 \\
               & $\{1, 1/2, 1/4, 1/8\}$ & 94.2 & 96.4 & 96.5 & 98.8 & 95.4 & 97.6 \\
        \midrule
                & $\{1, 1/2\}$            & 93.9 & 95.8 & 95.4 & 97.7 & 94.7 & 96.7 \\
          $l=8$ & $\{1, 1/2, 1/4\}$       & 94.0 & 96.5 & 95.8 & 99.1 & 94.9 & 97.8 \\
                & $\{1, 1/2, 1/4, 1/8\}$ & 94.4 & 96.0 & 95.8 & 98.1 & 95.1 & 97.0 \\
        \bottomrule
        \end{tabular}
\end{wraptable}

\paragraph{Impact of patch size and scale factors}
We further examined the effects of patch size and resampling scales by testing different hyperparameter settings, as shown in Table~\ref{evaluation table on different hyperparameters}. The best performance was achieved with the smallest patch size ($l=2$), indicating that stronger semantic scrambling improves detection. Moderate multi-scale fusion ($\mathbb{S} = \{1, 1/2, 1/4\}$) also led to optimal results, confirming the benefit of incorporating resampling artifacts.

\begin{wraptable}[8]{r}{0.45\textwidth}
    \caption{Impact of interpolation method.}
    \label{evaluation table on different methods of sampling}
    \centering
    \tabcolsep=0.3em
    \begin{tabular}{lrrrrrr}
        \toprule
        \multirow{2}{*}{Interp.} & \multicolumn{2}{c}{GAN-set} & \multicolumn{2}{c}{Diff.-set} & \multicolumn{2}{c}{Mean} \\
        & Acc. & A.P. & Acc. & A.P. & Acc. & A.P. \\
        \midrule
        \rowcolor{gray! 30} 
        Bilinear
         & \textbf{96.4} & \textbf{98.2} & 97.8 & \textbf{99.6} & \textbf{97.1} & 98.9 \\
         Bicubic
         & 96.2 & 97.9 & \textbf{97.9} & 99.3 & 96.9 & \textbf{99.1} \\
        Nearest
         & 94.6 & 97.4 & 96.8 & 99.2 & 95.7 & 98.3 \\
        \bottomrule
    \end{tabular}
\end{wraptable}

\paragraph{Impact of the resampling interpolation method}
We also evaluated the impact of interpolation methods, comparing bilinear, bicubic, and nearest-neighbor (Table~\ref{evaluation table on different methods of sampling}). Bilinear outperforms nearest-neighbor and performs comparably to bicubic. This might be because bilinear and bicubic blend neighboring pixel values, introducing richer entropy variations, while nearest-neighbor simply copies pixel values, resulting in limited entropy diversity.

\begin{wraptable}[8]{r}{0.45\textwidth}
    \caption{Impact of sliding window stride.}
    \label{tab: inter and intra textures}
    \centering
    \tabcolsep=0.35em
    \begin{tabular}{crrrrrr}
        \toprule
        \multirow{2}{*}{Stride} & \multicolumn{2}{c}{GAN-set} & \multicolumn{2}{c}{Diff.-set} & \multicolumn{2}{c}{Mean} \\
         & Acc. & A.P. & Acc. & A.P. & Acc. & A.P. \\
         \midrule
         \rowcolor{gray! 30} 
        1 & \textbf{96.4} & \textbf{98.2} & \textbf{97.8} & \textbf{99.6} & \textbf{97.1} & \textbf{98.9} \\
        2 & 94.9 & 97.3 & 97.0 & 99.5 & 95.9 & 98.4 \\
        \bottomrule
    \end{tabular}
\end{wraptable}

\paragraph{Impact of sliding window stride}
We evaluated the effect of stride in the $2\times2$ sliding window for LEP computation, testing strides of 1 and 2 (Table~\ref{tab: inter and intra textures}). A stride of 1 significantly outperforms 2, highlighting the importance of inter-block entropy in MLEP. This supports our multi-granularity design, which captures both intra- and inter-block texture patterns as depicted in Fig.~\ref{fig:entropy_illustration}.

\begin{wraptable}[7]{r}{0.5\textwidth}
    \caption{Evaluation on various ResNets.}
    \label{tab: different backbones}
    \centering
        \tabcolsep=0.3em
        \begin{tabular}{lrrrrrr}
        \toprule
        \multirow{2}{*}{Backbone} & \multicolumn{2}{c}{GAN-set} & \multicolumn{2}{c}{Diff.-set} & \multicolumn{2}{c}{Mean} \\
         & Acc. & A.P. & Acc. & A.P. & Acc. & A.P. \\
         \midrule
        ResNet-18  & 96.0 & 97.9 & 97.7 & 99.5 & 96.8 & 98.7 \\
        ResNet-34  & 96.1 & 98.2 & 97.7 & \textbf{99.7} & 96.9 & 98.9 \\
        ResNet-50  & 96.4 & 98.2 & 97.8 & 99.6 & 97.1 & 98.9 \\
        ResNet-101 & \textbf{96.4} & \textbf{98.3} & \textbf{97.8} & 99.6 & \textbf{97.1} & \textbf{99.0} \\
        \bottomrule
        \end{tabular}
\end{wraptable}

\paragraph{Compatibility with various backbones}
Lastly, we assessed the compatibility of MLEP with various ResNet backbones~\cite{he2016resnet}, including ResNet-18, 34, 50, and 101. As shown in Table~\ref{tab: different backbones}, all variants achieved strong performance, with slight gains from larger models. This confirms the generality and scalability of the proposed feature extraction method.

\subsubsection{Interpretability of MLEP}
To illustrate the effectiveness of MLEP for AIGI detection, 
we conducted a set of qualitative analysis detailed as follows.

\newcommand{\hscale}{0.35}
\begin{figure}[t]
  \centering
  \subfloat{
    \includegraphics[height=\hscale\textwidth]{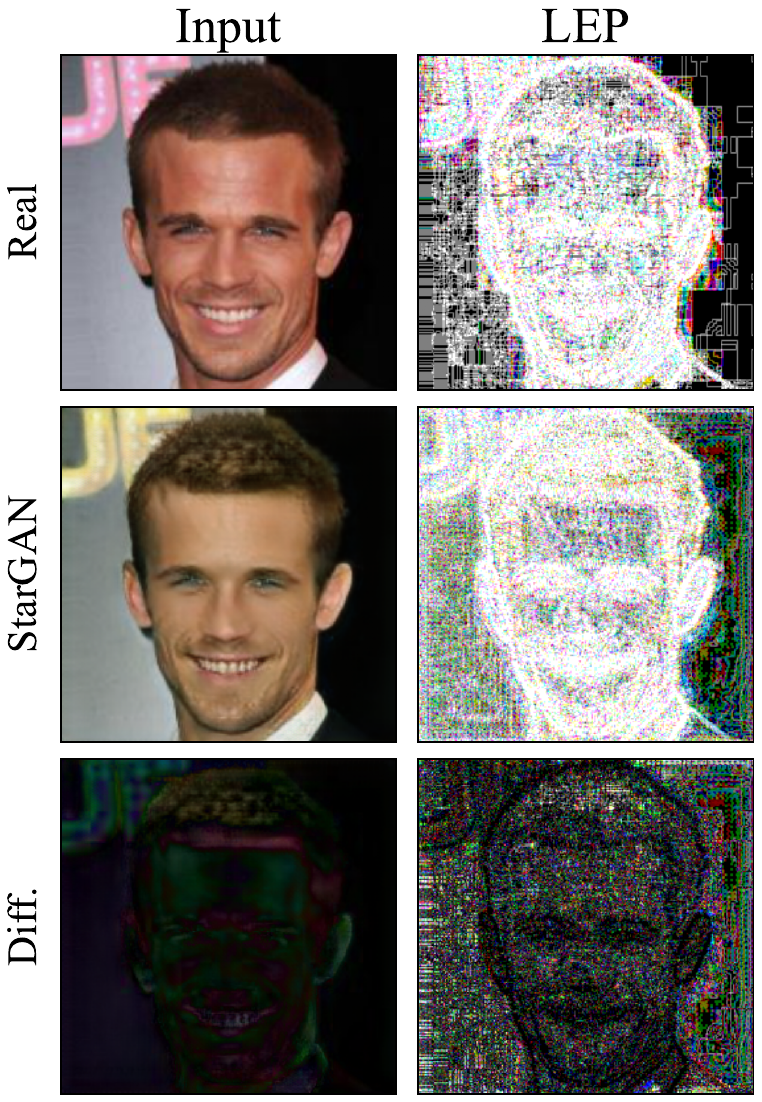}
  }
  \subfloat{
    \includegraphics[height=\hscale\textwidth]{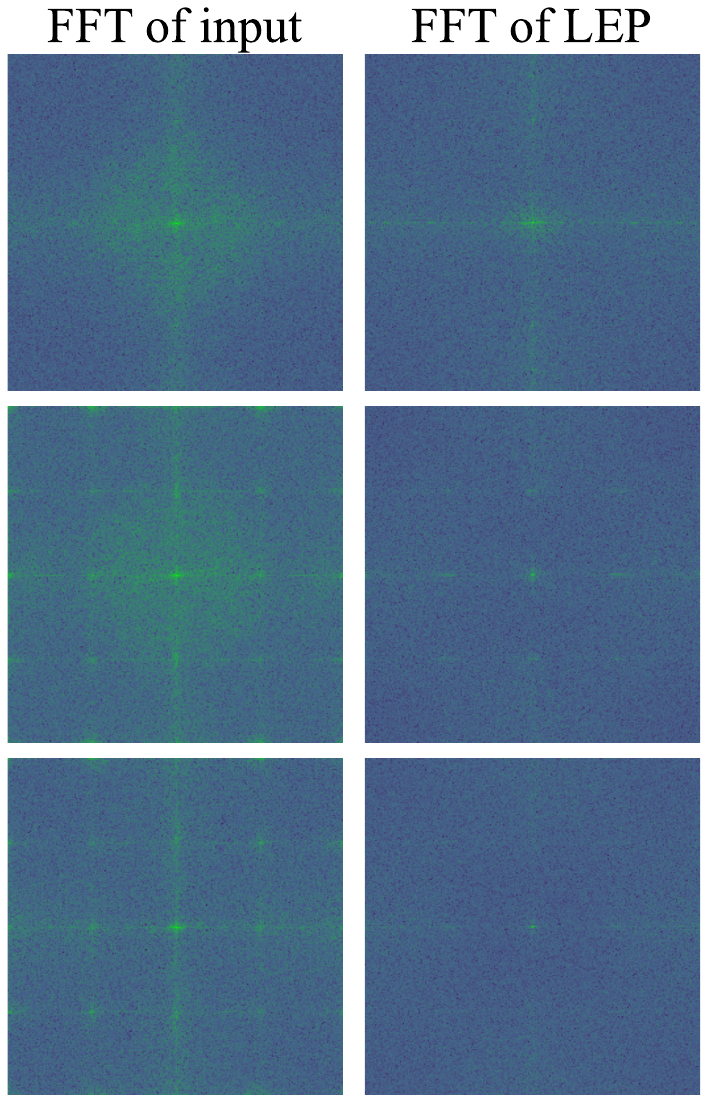}
  }
  \subfloat{
    \includegraphics[height=\hscale\textwidth]{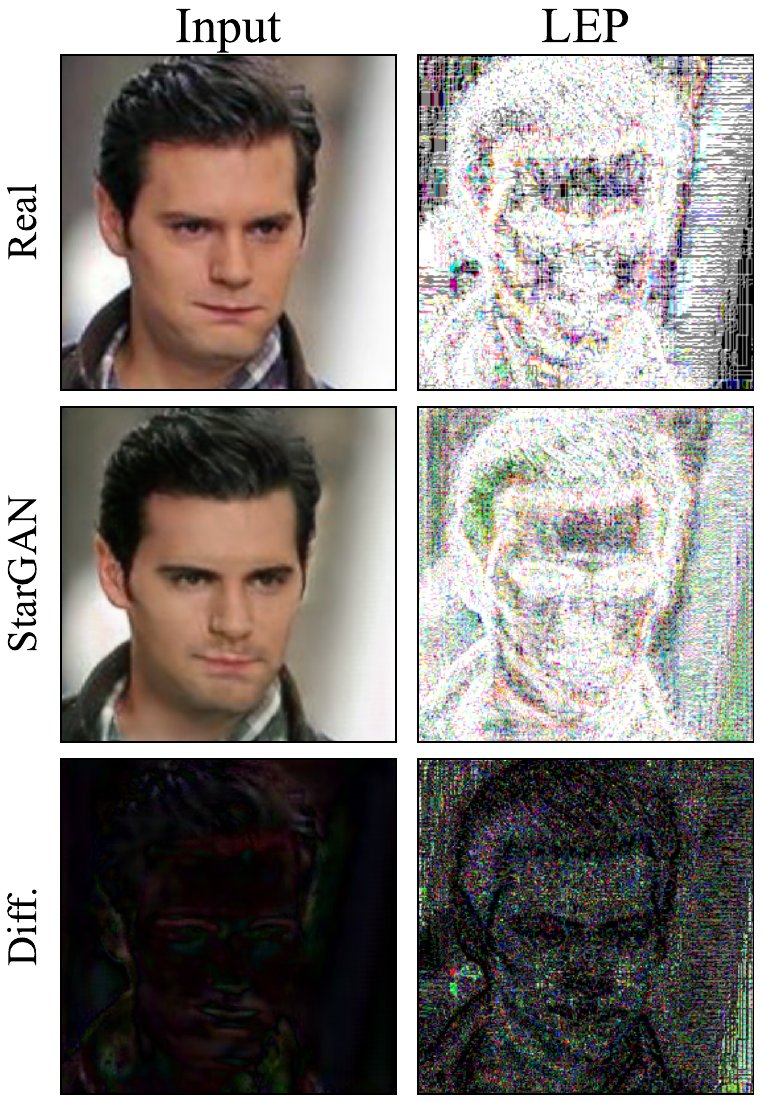}
  }
  \subfloat{
    \includegraphics[height=\hscale\textwidth]{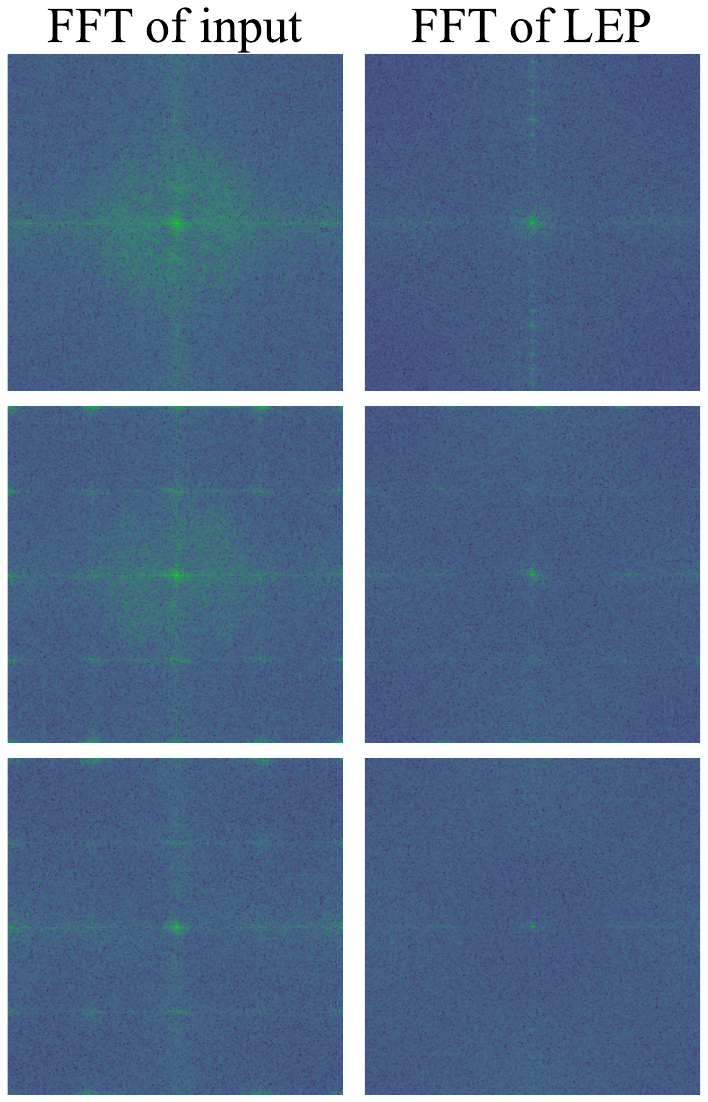}
  }\\
  \subfloat{
    \includegraphics[height=\hscale\textwidth]{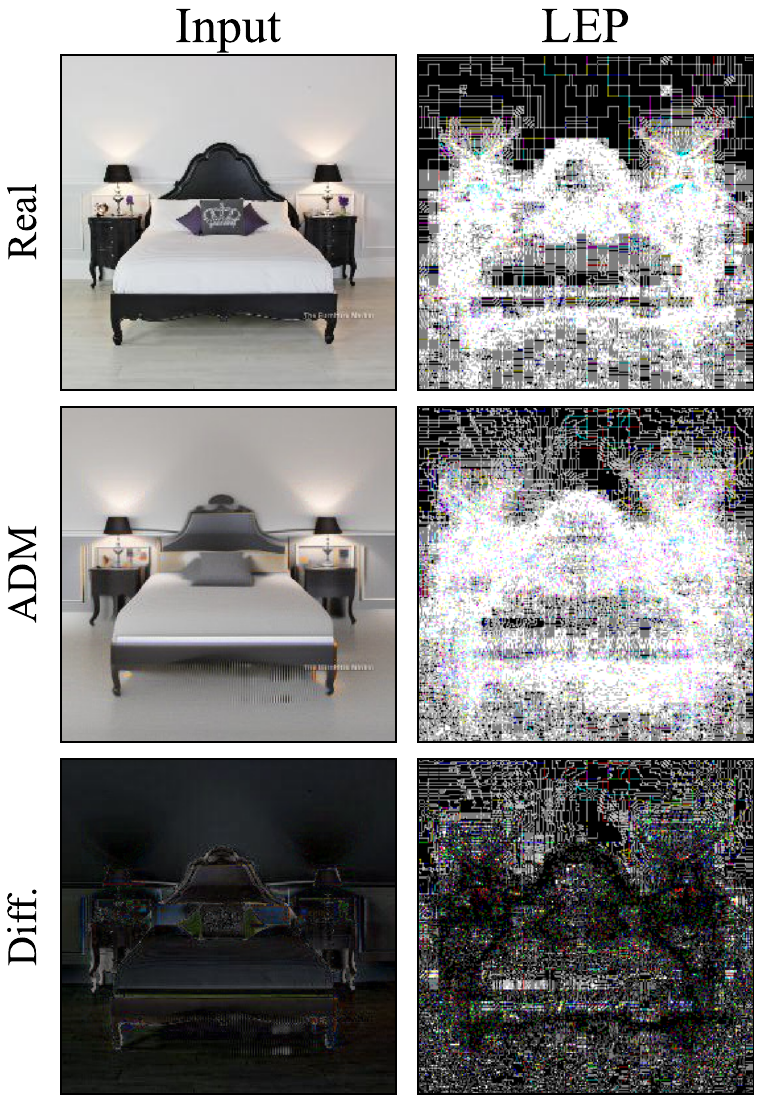}
  }
  \subfloat{
    \includegraphics[height=\hscale\textwidth]{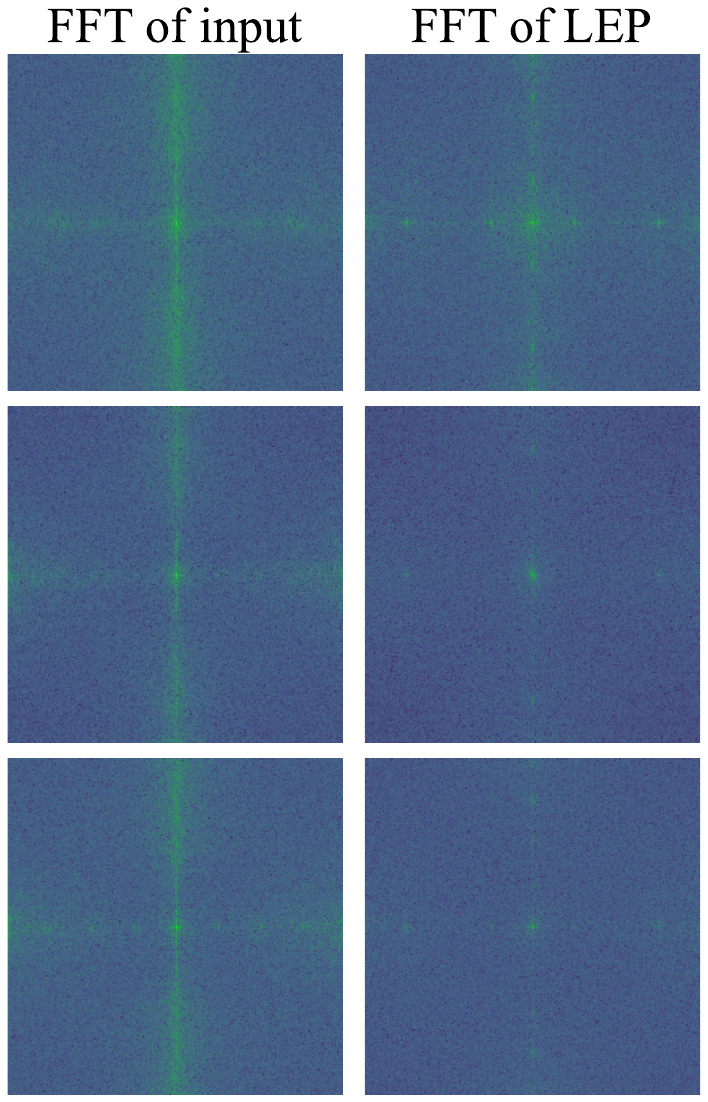}
  }
  \subfloat{
    \includegraphics[height=\hscale\textwidth]{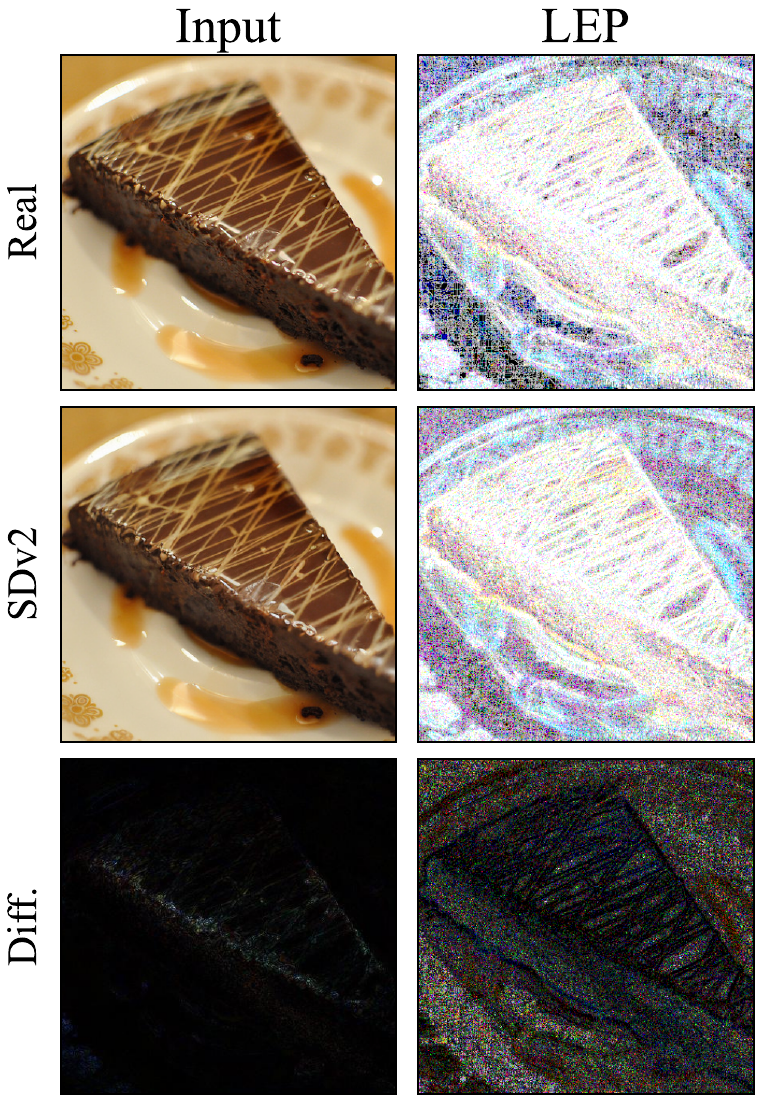}
  }
  \subfloat{
    \includegraphics[height=\hscale\textwidth]{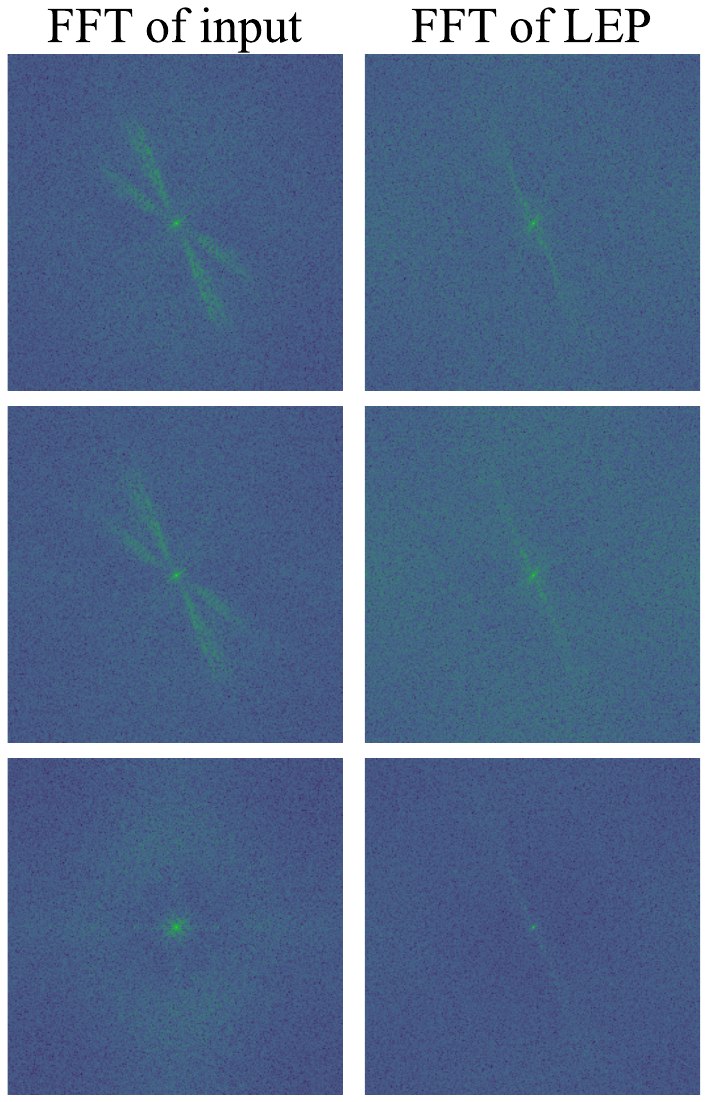}
  }
  \caption{Visualization of local entropy patterns for several real–fake image pairs, along with their differences in the pixel, entropy, and Fourier domains.}
  \label{fig:plot_lep_diff}
\end{figure}

\paragraph{Entropy patterns between real and AI-generated images}
We first visualize LEP maps for several real–fake image pairs, along with their differences in the pixel, entropy, and Fourier domains. Here, ``fake'' refers to AI-reconstructed images resembling the originals. Since LEP values are sparse and capped below 2.0, we normalize them to $[0, 255]$ for visualization. As shown in Fig.~\ref{fig:plot_lep_diff}, LEP differences are far more pronounced than pixel-level differences, especially for high-quality generations like Stable Diffusion v2, where pixel differences are visually negligible. In the frequency domain, real–fake differences show more consistent patterns than in the pixel space, supporting content-agnostic detection. These results highlight LEP's ability to amplify real–fake discrepancies while minimizing semantic interference.

\begin{figure}[t]
    \centering
    \includegraphics[width=\columnwidth]{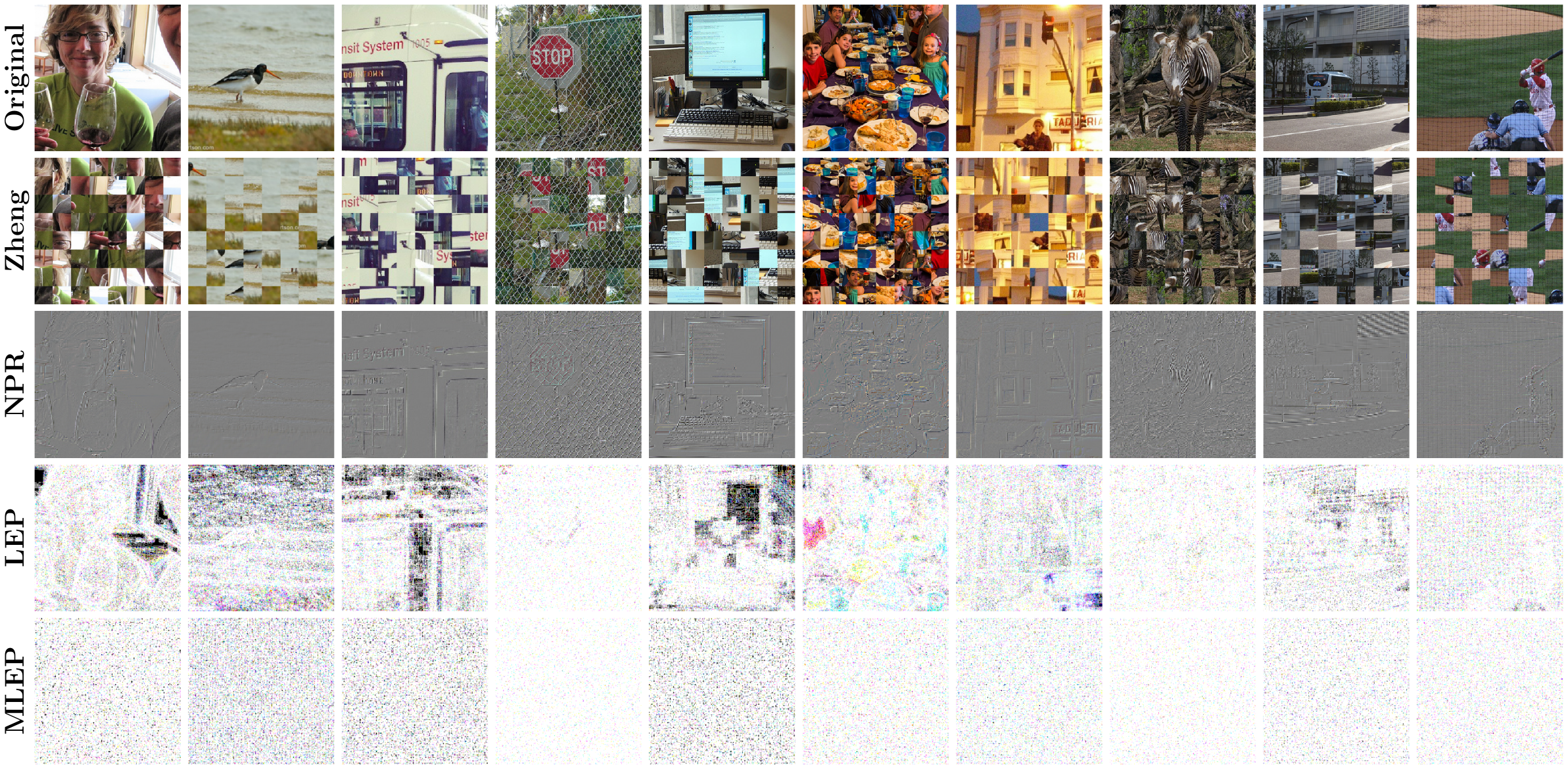}
    \caption{Qualitative comparison among Zheng~\cite{zheng2024breaking}, NPR~\cite{tan2024rethinking}, and our method. LEP preserves minimal visible semantics, while MLEP (without resampling) further suppresses semantic content.}
    \label{fig:plot_npr_compare}
\end{figure}

\paragraph{Semantic suppression capability of MLEP}
We further examine the semantic suppression capability of MLEP compared to two competitive methods: Zheng~\cite{zheng2024breaking} and NPR~\cite{tan2024rethinking}. Fig.\ref{fig:plot_npr_compare} visualizes feature maps of LEP and MLEP (without multi-scale resampling), alongside those from Zheng~\cite{zheng2024breaking} and NPR~\cite{tan2024rethinking}. The $32 \times 32$ shuffled patches in Zheng\cite{zheng2024breaking} still retain noticeable semantic cues both locally and globally. NPR~\cite{tan2024rethinking} produces edge-like features by computing pixel differences, leaving much of the original semantics intact. In contrast, LEP substantially suppresses semantic content by highlighting pixel-level randomness, and MLEP further eliminates it through fine-grained patch shuffling, enabling learning content-agnostic representation for AIGI detection.

\begin{wrapfigure}[27]{r}{0.65\textwidth}
    \centering
    \includegraphics[width=0.65\textwidth]{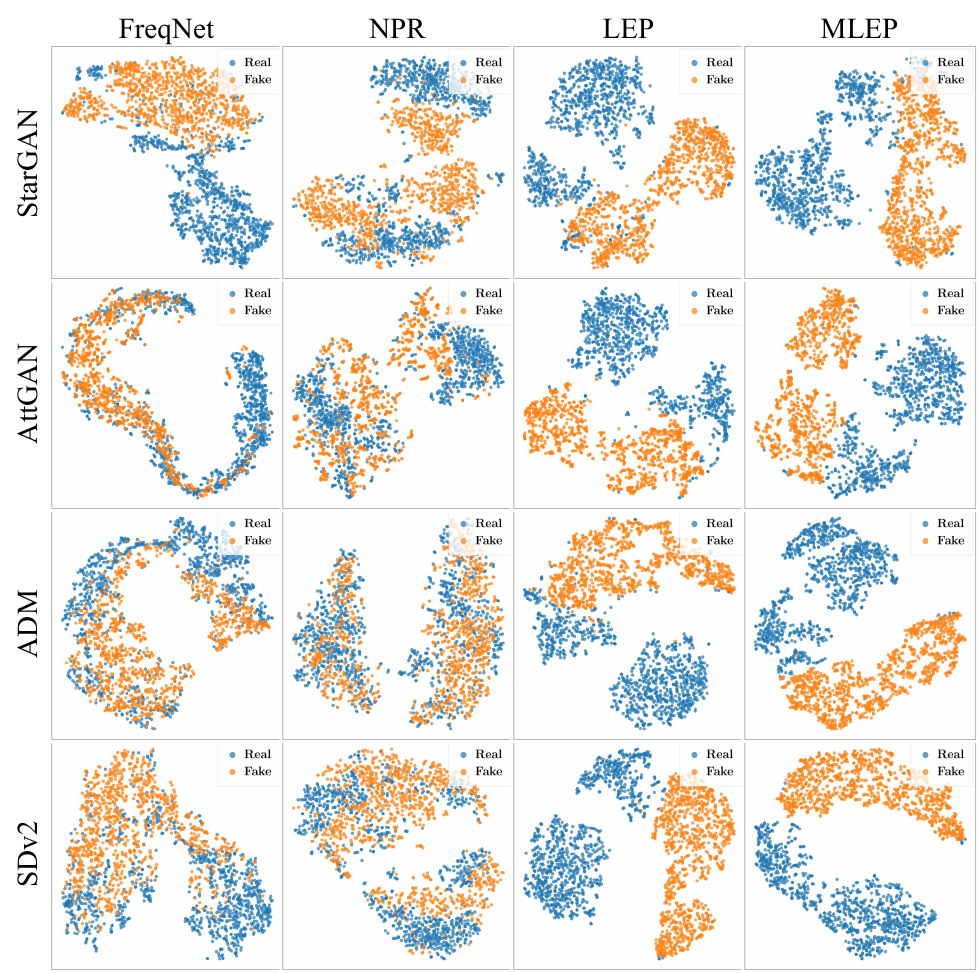}
    \caption{t-SNE visualization of real vs. fake samples across methods.}
    \label{fig:tSNE_plot}
\end{wrapfigure}

\paragraph{Feature distribution of real and AI-generated images} 
Finally, Fig.\ref{fig:tSNE_plot} visualizes the t-SNE distribution~\cite{van2008visualizing} of real and fake samples based on the final feature layer of a ResNet-50 classifier, comparing our method with two competitive baselines—NPR~\cite{tan2024rethinking} and FreqNet~\cite{tan2024frequency}—which also use ResNet-50. We showcase results on four generative models: StarGAN~\cite{choi2018stargan}, AttGAN~\cite{he2019attgan}, ADM~\cite{dhariwal2021diffusion}, and SDv2~\cite{rombach2022high}. The proposed local entropy patterns (LEP) achieve noticeably cleaner real–fake separation than the baselines, and MLEP further enhances this distinction, demonstrating stronger discriminative capability for AIGI detection.


\section{Conclusion} \label{sec:conclusion}

This paper explores the use of image entropy as a cue for detecting AI-generated images (AIGI) and introduces Multi-granularity Local Entropy Patterns (MLEP), a set of entropy-based feature maps derived from shuffled small patches across multiple image scales. MLEP captures pixel relationships across spatial and scale dimensions while disrupting image semantics, thereby mitigating content bias. Based on MLEP, we train a ResNet-based classifier that achieves robust detection performance. Experiments on 32 generative models show that our method outperforms state-of-the-art approaches by over 3\% in accuracy.

Nonetheless, limitations remain. The entropy in this work is computed within $2 \times 2$ windows only due to the exponential increase in complexity with larger windows. Additionally, the robustness of entropy patterns under common image processing operations (e.g., compression, blurring, and noise addition) warrants further study. Future work may explore more efficient and adaptive entropy computation methods to enhance detection performance.



\bibliographystyle{plain}
\bibliography{sections/references}

\end{document}